\documentclass{article}

\usepackage{microtype}
\usepackage{graphicx}
\usepackage{subfigure}
\usepackage{booktabs} 
\usepackage{caption}

\usepackage[hidelinks]{hyperref}

\usepackage{PRIMEarxiv}
\usepackage{natbib}
\usepackage{xcolor}

\usepackage{amsmath}
\usepackage{amssymb}
\usepackage{mathtools}
\usepackage{amsthm}

\usepackage[capitalize,noabbrev]{cleveref}

\usepackage{amsmath,amsfonts,bm}

\def\eqref#1{equation~\ref{#1}}

\def\1{\bm{1}}

\DeclareMathAlphabet{\mathsfit}{\encodingdefault}{\sfdefault}{m}{sl}
\SetMathAlphabet{\mathsfit}{bold}{\encodingdefault}{\sfdefault}{bx}{n}

\DeclareMathOperator*{\argmax}{arg\,max}
\DeclareMathOperator*{\argmin}{arg\,min}

\usepackage{adjustbox}
\usepackage[page]{appendix}

\usepackage{colortbl}
\usepackage{multirow}
\newcommand\myeq{\mathrel{\stackrel{\makebox[0pt]{\mbox{\normalfont\tiny def}}}{=}}}
\theoremstyle{plain}
\newtheorem{theorem}{Theorem}[section]

\theoremstyle{definition}
\newtheorem{definition}[theorem]{Definition}

\theoremstyle{remark}

\title{
Semi-Supervised Learning of Pushforwards for Domain Translation \& Adaptation 
\thanks{\textit{\underline{Citation}}: 
\textbf{Authors. Title. Pages.... DOI:000000/11111.}} 
}
\author{
  Nishant Panda, Natalie Klein \\
  Los Alamos National Lab \\
  Los Alamos\\
  \texttt{\{nishpan, neklein\}@lanl.gov} \\
   \And
  Dominic Yang \\
  Department of Mathematics \\
  University of Californa, Los Angeles \\
  Los Angeles\\
  \texttt{domyang@g.ucla.edu} \\
  \AND
  Patrick Gasda \\
  Los Alamos National Lab\\
  \texttt{gasda@lanl.gov} \\
  \And
  Diane Oyen \\
  Los Alamos National Lab \\
  \texttt{doyen@lanl.gov} \\
}
\begin{document}
\maketitle

\begin{abstract}
Given two probability densities on related data spaces, we seek a map pushing one density to the other while satisfying application-dependent constraints.
For maps to have utility in a broad application space (including domain translation, domain adaptation, and generative modeling), the map must be available to apply on out-of-sample data points and should correspond to a probabilistic model over the two spaces. 
Unfortunately, existing approaches, which are primarily based on optimal transport, do not address these needs.
In this paper, we introduce a novel pushforward map learning algorithm that utilizes normalizing flows to parameterize the map. 
We first re-formulate the classical optimal transport problem to be map-focused and propose a learning algorithm to select from all possible maps under the constraint that the map minimizes a probability distance and application-specific regularizers; thus, our method can be seen as solving a modified optimal transport problem. 
Once the map is learned, it can be used to map samples from a source domain to a target domain.
In addition, because the map is parameterized as a composition of normalizing flows, it models the empirical distributions over the two data spaces and allows both sampling and likelihood evaluation for both data sets. 
We compare our method (parOT) to related optimal transport approaches in the context of domain adaptation and domain translation on benchmark data sets.
Finally, to illustrate the impact of our work on applied problems, we apply parOT to a real scientific application: spectral calibration for high-dimensional measurements from two vastly different environments.

\end{abstract}

\section{Introduction}\label{sec:intro}
The current landscape in data analysis is rich in heterogeneous data acquired from multiple sensors, experiments, and other data generation procedures. 
While large amounts of data enable data-driven algorithms, the problem of inference in the presence of such heterogeneous data still remains. 
For example, in predictive modeling, the usual training-testing assumption is that new samples will come from the same distribution as the training data; when the goal is to deploy a trained model under violations of this assumption, we need to account for discrepancies in data distribution. 
For instance, in scientific experiments, the same physical phenomenon may be observed by two different sensors, or, in certain situations, physics-based simulations may be used to supplement expensive experiments. 
In fields as disparate as remote sensing, speech recognition, and image analysis, such data discrepancies may lead to erroneous conclusions from data-driven models. 
In recent years, optimal transport (OT), a class of mathematical tools for distribution alignment, has gained increased attention in the machine learning community. 
In OT, the goal is to align two distributions by minimizing a transport distance over training samples. 
Here, the basic assumption is that the two data spaces with different distributions are connected via a \textit{pushforward map}, and the goal is to learn this map by doing  minimal work when moving a point from one data space to the other. 
In general, the constraint that the map be the pushforward is difficult to apply; in practice, almost all practical OT algorithms rely on a minimization algorithm over a suitable class of coupling functions describing all possible joint distributions on the two data spaces. 
Importantly, this procedure only provides an alignment over training data points from both domains. 
However, many practical cases demand more than an algorithm for aligning training data points. In many scientific applications, including sensor calibration, the following tasks are needed:
\begin{itemize}
\item \textbf{Task 1 (transformation)}: out-of-sample data transformation. 
\item \textbf{Task 2 (domain knowledge)}: incorporation of domain knowledge via regularization. 
\item \textbf{Task 3 (probabilistic model)}: abilities such as likelihood estimation and data generation (sampling).
\end{itemize}

To further illustrate the problem, we consider an example scientific application: the ChemCam instrument on board the Mars rover Curiosity.
ChemCam is part of the Mars Science Laboratory, a suite of instruments designed to study geologic and biological markers on Mars; findings could reveal factors and impacts of drastic climate change and the possibility of life on Mars.
ChemCam measures laser-induced breakdown spectroscopy (LIBS), a type of atomic emission spectroscopy in which a laser creates a small plasma spot on the target surface, causing the atoms to emit light that is collected by high-resolution spectrometers.
Since landing at Gale Crater in 2012, the ChemCam 
instrument has obtained spectral measurements of thousands of Martian rock and soil analysis targets~\citep{maurice2016chemcam}. 
The primary driver in variation across spectra is the chemical composition of the target, so previous works seek to predict the composition from a spectrum using methods like linear regression, dimension reduction, and deep neural networks  \citep[\textit{inter alios}]{wiens2013pre,forni2013independent, clegg2017recalibration, anderson2017improved, castorena2021deep}.
However, because the Martian targets have unknown chemical compositions, supervised learning approaches are, out of necessity, trained on \textit{calibration data} collected from laboratory instrument on Earth.
Exactly how differences in the Earth and Mars instruments and environments affect the spectra (and the resulting supervised learning models), and how to correct such differences, is an active area of research~\citep{anderson2022post}.
We seek to develop an Earth-to-Mars transform (task 1), in part to improve supervised learning models that are trained on Earth calibration spectra and applied to Mars spectra. 
In addition, the map should incorporate available domain knowledge (task 2), and a probabilistic model can provide principled methods for other uses such as outlier detection and sampling (task 3).

To address problems such as the ChemCam sensor calibration problem, we introduce a novel pushforward estimation algorithm (parOT) that aims to solve the three tasks using normalizing flows (NFs). 
Classically, OT frameworks are based on the Kantorovich relaxation of the Monge method in which the pushforward constraint is formulated as a minimization over a coupling matrix that describes the mapping over the empirical distributions of the training data. 
However, the practical realization of this formulation makes it difficult to complete task 1 (transformation). 
While some work addresses task 1 by learning a mapping along with the coupling matrix~\citep{perrot2016mapping} or by learning to interpolate a coupling matrix (e.g.,~\citealt{manole2021plugin}), it is often difficult to incorporate domain knowledge, and likelihood estimation or sampling of new data points may not be directly available.  
Other recent work partially addresses task 2 by modifying classical OT to include a certain form of map constraint~\cite{gukeypoint}; however, accomplishing task 1 in this framework requires non-trivial modifications, while task 3 (probabilistic model) is again not available. 
Moreover, as the map constraints become more complicated (e.g., physics based constraints), it is not clear how to include them in the learning algorithm. 

To remedy these shortcomings, we first re-formulate the classical OT problem of constrained optimization by making it \emph{map}-focused (i.e., we select from all possible maps under the constraint that the map minimizes a probability distance, formulated as an integral measure, such as the Wasserstein distance). Further, the map is parameterized as a composition of NFs that matches the empirical distribution of the two data spaces and makes use of a simpler base density for sampling and likelihood estimation (similar to~\citealt{marzouk2016introduction}). 
In Section~\ref{sec:rel_work}, we introduce notation and briefly discuss OT methods, NFs, and other related work, and in Section~\ref{sec:pushforward}, we describe our framework. 
Then, we demonstrate the novelty and efficacy of our approach compared to related methods on synthetic benchmark data sets in which data from a source domain is transported to a target domain via linear and non-linear transformations (Section~\ref{sec:results_benchmark}). 
Finally, we apply our framework to the ChemCam scientific application, learning an Earth-to-Mars transform for domain translation, adaptation, and likelihood estimation (Section~\ref{sec:e2m_dataset}).

\section{Related Work}
\label{sec:rel_work}
In this section, we introduce notation and give more detail and references for related work.
Let $\Omega_s,\Omega_t$, be the spaces describing the source and target inputs, respectively. In this paper, we will consider input spaces as finite-dimensional Euclidean spaces and thus assume that $\Omega_s\subset \mathbb{R}^{d_s}$ and $\Omega_t \subset \mathbb{R}^{d_t}$, respectively, for integers $d_s,d_t$. We further assume that $\mu_s,\mu_t$ are probability measures associated with the (sigma algebra of ) source and target spaces, respectively.
\begin{definition}\label{def:pushforward}
A measurable function $F: \Omega_s\to \Omega_t$ induces a measure on $\Omega_t$ called the \textit{pushforward measure}~\citep[Section $3.6$, under Image of Measures]{bogachev2007measure} given by
\[{F}\sharp\mu_{s}[\cdot] = \mu_{s}({F}^{-1}[\cdot]),\]
where ${F}^{-1}$ is the pre-image (inverse image) and $[\cdot]$ is any measurable subset of $\Omega_t$. 
\end{definition}


\subsection{Optimal Transport (OT)}
OT for machine learning domain adaptation was introduced nearly a decade ago~\citep{courty2014domain}, with modern works based on OT indicating continued interest in the machine learning community~\citep[\textit{inter alios}]{courty2017joint, damodaran2018deepjdot, balaji2020robust, fatras2021unbalanced}.
The starting point of the OT framework is given by the following minimization problem due to Monge~\citep{santambrogio2015optimal},
\begin{equation}
\begin{small}
\label{eq:monge}
    T_0 = \underset{T:\Omega_s\to \Omega_t}{\argmin} \int_{\Omega_s}c(T(x),x)d\mu_s(x)\quad \text{s.t. \, } {T}\sharp\mu_{s} =  \mu_t,
\end{small}
\end{equation}
where $c$ is a cost function. In practice, it is difficult to apply the constraint to the map $T$ directly, and, moreover, the constraint is not closed under weak convergence. To alleviate this issue, the standard regularized OT Kantorovich formulation of the Monge problem (Equation~\ref{eq:monge}), developed by~\citealt{courty2014domain}, seeks a coupling matrix $\gamma$ satisfying
\begin{align}
    \hat\gamma = \arg \min_{\gamma \in \mathcal{P}} \langle \gamma, C \rangle_F + \lambda h(\gamma),
\end{align}
where $\mathcal{P}$ is the space of joint distributions over source and target, $C$ is a cost matrix computed between training source and target samples, $\langle \cdot \rangle_F$ is the Frobenius norm, and $h$ is an entropy regularization term. This formulation allows computationally-efficient algorithms to solve for $\gamma$.

However, $\gamma$ only couples the source and target samples used to solve the OT problem, making it difficult to apply the learned transformation to new samples. Some attempts have been made to learn a closed-form map function during solution of the OT problem (via alternating optimization); specifically,~\citealt{perrot2016mapping} posits linear or kernel-based maps. 
Even if these maps are suitably flexible, it is still difficult to include map constraints.
Paired data constraints were very recently addressed in the context of OT under the name keypoint-guided OT (KGOT)~\citep{gukeypoint}.
KGOT involves i) a mask applied to $\gamma$ and ii) an additional regularization term based on the distances between other points and the paired points.
While KGOT includes paired data constraints, it is unclear how to extend it to other constraints, and does not include map estimation without non-trivial extensions. In summary, achieving both tasks 1 and 2  is highly non-trivial in current OT methods, and to our knowledge, task 3 (probabilistic model) is not directly addressed by any existing OT methods.

\subsection{Normalizing Flows (NFs)}
NFs are a flexible class of neural networks parameterizing diffeomorphic maps on Euclidean spaces that are used for learning complex probability distributions with good universal approximation properties~\citep{lee2021universal,shah2022learning}.
That is, NFs model the data probability density function $p_X(x)$ through a function $f_\phi$ applied to random variates from a simpler density function $p_Z(z)$ (see~\citealt{papamakarios2021normalizing} for a recent review). 
Using the change of variables formula, the data likelihood can be expressed:
\begin{align}
    p_X(x) = p_Z(f_\phi(x)) \left| \frac{\partial f_\phi(x)}{\partial x} \right|.
\end{align}
Normalizing flows are trained by minimizing the negative log likelihood (NLL) with respect to parameters $\phi$ via stochastic gradient descent.
However, the function $f_\phi$ must be invertible, and for computational efficiency, both the inverse $f_\phi^{-1}$ and Jacobian determinant should be simple to calculate. 
Modern normalizing flows with these properties that we utilize in this paper include RealNVP~\citep{dinh2016density} and neural spline flows~\citep{durkan2019neural}.

\subsection{Other Related Approaches}
Our work differs from previous work in domain adaptation that seeks to learn a feature space invariant to domain (e.g.,~\citealt{ganin2015unsupervised}); we do not learn an invariant feature space.
In addition, unlike domain adaptation-focused works, our framework addresses all three tasks listed in Section~\ref{sec:intro}, with domain adaptation a potential application of the map learned in task 1 (transformation) rather than the primary end goal.
Other related approaches tend to focus on task 1.
Similar to our Triangle NF (defined in Section~\ref{sec:pushforward}), AlignFlow~\citep{grover2020alignflow} uses composed NFs to learn a cycle-consistent map between domains, but the loss function (a combination of maximum likelihood and adversarial losses) differs significantly from ours, as it does not directly include probability distance. Furthermore, any constraints on the map are enforced indirectly by sharing weights across two NFs, complicating straightforward incorporation of domain knowledge. 
Despite these differences, our framework does share some properties (such as cycle consistency) with AlignFlow.
Another work concurrent to ours utilizes normalizing flows to learn an OT-like map with sliced Wasserstein distance \citep{coeurdoux2022sliced}.
While the map construction (addressing Task 1) bears similarities to our method, there is no tractable base density, limiting the use of this method for likelihood evaluation or sampling (task 3), and the authors did not incorporate constraints (task 2).
Another group of approaches appeal to or seek to improve OT for domain translation.
The work of \citealt{de2019optimal} proposes learning a one-way mapping between two domains based on dynamical systems and relates the objective to OT;
\citealt{li2020enhanced} is based on improvements to OT through an attention mechanism, a neural network Kantorovich potential (similar to~\cite{makkuva2020optimal}), and inclusion of the source classification loss into the learning objective;
\citealt{lu2019guiding} seek to improve CycleGAN~\citep{zhu2017unpaired} by regularizing with OT barycenters, which incorporates the OT objective indirectly; and \citealt{seguy2017large} takes a two-step approach, in which the standard OT problem is solved, then a parameterized map is learned based on the OT barycenters.
More recent works in this direction improve on computational methods and provide statistical guarantees for such out-of-sample OT maps \citep{muzellec2021near,manole2021plugin,hutter2021minimax,pooladian2021entropic}.
In another line of work, authors seek to improve properties of normalizing flows with OT loss functions \citep{huang2020convex,morel2022turning}.
While these works may improve OT, they do not directly address tasks 2 and 3.
Our method utilizes a modified primal Monge formulation with constraints and an integral probability metric Lagrangian. Some recent related works instead work with the dual Monge formulation~\citep{korotin2021neural,rout2021generative,asadulaev2022neural}. Another work that uses the primal formulation needs non-trivial extensions to incorporate constraints, and while sampling is available, it is not well suited for Task 3~\citep{fan2021scalable}.
Table~\ref{tab:related_work_tasks} provides a summary of related methods in terms of ability to solve tasks 1, 2, and 3.

\begin{table*}[]
    \centering
    \caption{Summary of selected related methods and whether they natively accomplish tasks 1, 2, and 3. Here, we separate out task 3 into two example tasks, sampling and likelihood evaluation, and mark X if a method easily allows this task. (Task 1: out of sample translation; task 2: incorporation of non-OT constraints on the map; task 3: probabilistic model.)(* indicates non-trivial adjustments to the framework)}
    \label{tab:related_work_tasks}
    \begin{tabular}{l|c|c|c|c}
        Method & Task 1 & Task 2 & Task 3a (sampling) & Task 3b (likelihood) \\ \hline
        parOT (ours) & X & X & X & X \\ 
        \citealt{pooladian2021entropic} & X & & X \\
        \citealt{manole2021plugin} & X & & X \\
        \citealt{hutter2021minimax} & X &&& \\
        \citealt{muzellec2021near} & X &&& \\
        \citealt{coeurdoux2022sliced} & X &&& \\
        \citealt{korotin2021neural} & X &&& \\
        \citealt{korotin2022neural} & X &&& \\
        \citealt{fan2021scalable} & X & X$^*$ & X & \\
        \citealt{asadulaev2022neural} & X & X$^*$ & X & \\
        \citealt{rout2021generative} & X & & X & \\
        \citealt{cuturi2013sinkhorn} & X & & & \\
    \end{tabular}

\end{table*}

\subsection{Benchmark comparison methods}
We compare to several related methods in our benchmark data experiments. 
In these experiments, a true pushforward map is known, and we measure how well each method recovers the pushforward map via pointwise mean squared error (MSE) on a set of paired test data.
The methods we compare to, including paper references and code attribution, are given in Table~\ref{tab:benchmark_methods}.

\begin{table*}[]
    \centering
    \caption{Methods we compare to in our benchmark experiments with paper references and code (* indicates that we modified existing code to provide a map estimate).}
    \label{tab:benchmark_methods}
    \begin{adjustbox}{width=\textwidth}
    \begin{tabular}{l|l|l}
        Name & Reference & Code \\ \hline
        Linear and Kernel OT & \citealt{flamary2021pot} & \url{https://github.com/PythonOT/POT} \\
        Linear and Kernel KGOT$^*$ & \citealt{gukeypoint} & \url{https://github.com/XJTU-XGU/KPG-RL} \\
        1NN & \citealt{manole2021plugin} & \url{https://github.com/APooladian/1NN_MapEstimator} \\
        SWOT-Flow & \citealt{coeurdoux2022sliced} & \url{https://github.com/FlorentinCDX/SWOT-Flow}\\
        Rout & \citealt{rout2021generative} & \url{https://github.com/LituRout/OptimalTransportModeling}\\
    \end{tabular}
    \end{adjustbox}

\end{table*}

Briefly, Linear and Kernel OT perform alternate optimization to both solve the standard OT problem and estimate a linear or kernel map that is faithful to the OT coupling.
KGOT solves the standard OT problem, but with constraints specifying that keypoints (paired data points) need to map to each other, meaning entries in the coupling matrix are set to 0 or 1 for those points; in addition, a regularization term is added to encourage points nearby the keypoints to map close to each other.
While KGOT does not provide an out-of-sample map, we modified the code to work similarly to Linear and Kernel OT to estimate a map.
1NN solves the standard OT problem then interpolates the coupling matrix with nearest neighbors.
SWOT-Flow uses normalizing flows to learn a map between two empirical distributions, but unlike our method, does not make use of a simple base density.
Rout uses the quadratic loss dual formulation of the Monge problem with maps specified as generative neural networks.
Unless otherwise specified in experiments, we used hyperparameters available in the relevant examples in each code base.

\section{Parametric Pushforward Estimation With Map Constraints (parOT)}
\label{sec:pushforward}

  We will assume that a (true) measurable function $F:\Omega_s \to \Omega_t$ exists such that the target distribution is the pushforward of the source distribution under $F$, i.e.,~$F\sharp\mu_{s} = \mu_t$. We don't know the exact form of $F$, but in certain applications (e.g., scientific applications), we can formulate well-defined constraints on $F$. While our framework allows flexible specification of map constraints, in this work, we focus on cases in which knowledge of the pushforward function $F$ is encoded by a \emph{paired} dataset $\mathcal{D}_p = \lbrace(x_k^s,x_k^t)\rbrace_{k = 1}^{N_p}$, where $F(x_k^s) = x_k^t$. We denote the joint distribution of source and target samples as $D_p$, so $\mathcal{D}_p \sim D_p$. 
 
 \subsection{Monge Formulation with Constraints}
 In the context of paired information constraints, we can formulate Equation~\ref{eq:monge} as the following constrained minimization problem
\begin{equation}
    T_0 = \underset{T:\Omega_s\to \Omega_t}{\argmin}
    H(T)
    \quad \text{s.t. } {T}\sharp\mu_{s} = \mu_t \myeq F\sharp\mu_{s} , 
\end{equation}
where 
{\small\begin{align*}
H(T) &= \left( \int_{\Omega_s} |T(x) - x|^p d\mu_s(x) \right)^{1/p}  \\
    &+ \left(\mathbb{E}_{(x^s,x^t)\sim D_{p}} \left\lvert T(x^s) - x^t\right\rvert^p \right)^{1/p},
\end{align*}}

In our approach, we \emph{lift} the pushforward constraint via an integral probability metric (IPM) $\gamma_{\mathcal{F}}(\mathbb{P},\mathbb{Q})$ (~\cref{def:IPM}). 
\begin{definition}\label{def:IPM}
\citep{sriperumbudur2009integral} Let $\mathbb{P},\mathbb{Q}$ be two probability measures defined on the same sigma algebra over a set $M$, the integral probability metric $\gamma_{\mathcal{F}}(\mathbb{P},\mathbb{Q})$ is given by
\begin{equation}\label{eq:IPM}
    \gamma_{\mathcal{F}}(\mathbb{P},\mathbb{Q}) = \underset{\phi\in\mathcal{F}}{\sup}{\left\lvert \int_{M}\phi d\mathbb{P} - \int_M\phi d\mathbb{Q}\right\rvert}
\end{equation}
where $\mathcal{F}$ is a suitable class of real valued measurable functions on $M$.

\end{definition}
Some examples of IPMs are the 1-Wasserstein distance $W^1$, Maximum Mean Discrepancy, or the total variation distance.
Our parametric OT framework can be formulated as the learning problem in~\cref{def:objective_function}.
\begin{definition}\label{def:objective_function}
The learning problem associated with our methodology is given by
\begin{equation}
\label{eq:param_OT_opt}
T_0 = \underset{T:\Omega_s\to \Omega_t}{\argmin} H(T) + \lambda \gamma_{\mathcal{F}}\left(\mu_t,{T}\sharp\mu_{s}\right),
\end{equation}
\end{definition}



In the following paragraphs, we will describe the learning algorithm for~\cref{eq:param_OT_opt} for the particular case when ${d_s} = d_t$. 
We will further assume that the true function $F$ is a diffeomorphism from the source to target (i.e., it is a bijective function with a differentiable inverse). With this assumption, if $\mu_s$ is represented by a probability density function, then the target measure, which is the pushforward under $F$ (i.e., $\mu_t = F\sharp\mu_s$) is given by the change of variables formula~\citep{folland1999real}. While this may seem very restrictive, in almost all cases such a restriction on the true function is not an obstacle. Indeed, the diffeomorphism is allowed to exist almost everywhere (a.e.) i.e.~modulo a null set. For the general case when $d_s \neq d_t$, we will describe adjustments to the framework in~\cref{sec:app_learning_algo} in the Appendix. We will refer to our framework as $\text{par}$OT. 

\subsection{Map Construction and Learning Algorithm}
\begin{figure}
    \vskip 0.2in
    \begin{center}
    \includegraphics[width=0.6\linewidth]{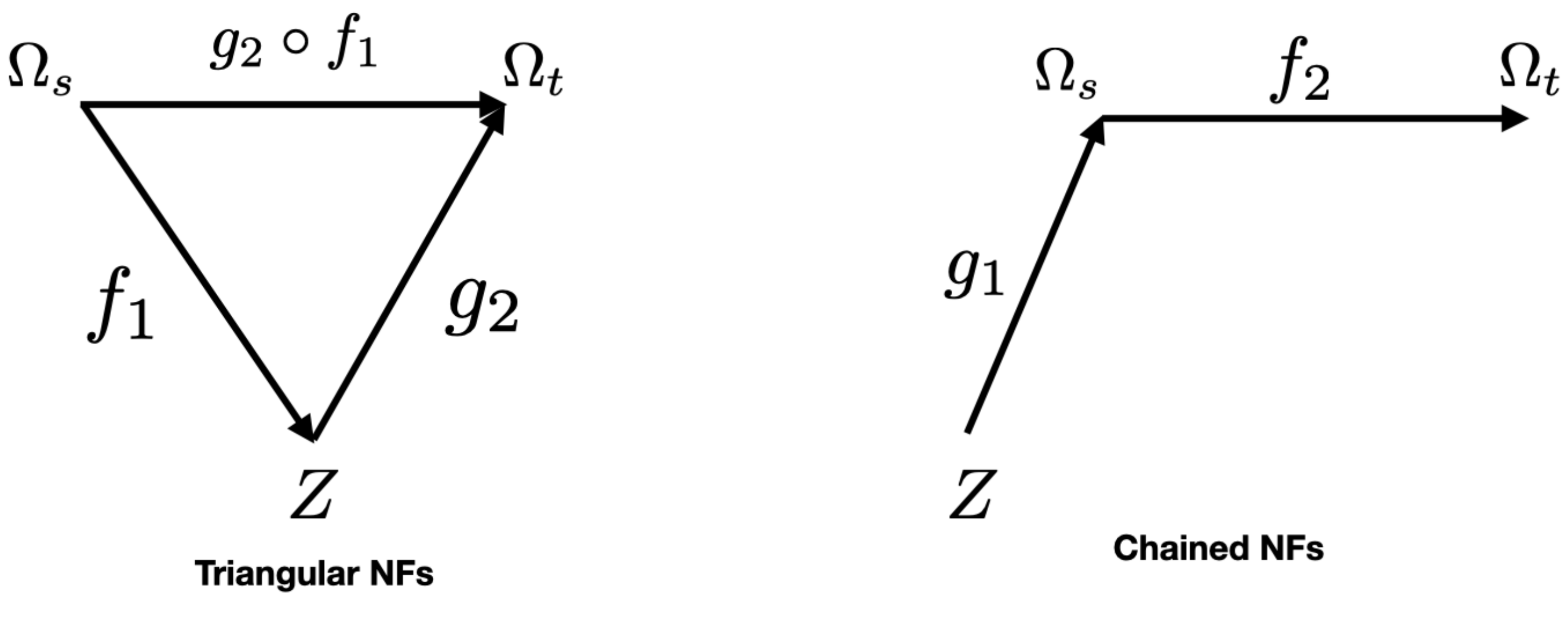}
    \caption{Two schemes for the construction of the source-to-target parOT pushforward map. On the left (Triangle NFs), we construct NFs $f_1,f_2$ (along with their inverses $g_1,g_2$) transforming the source and target distribution to the standard Gaussian on $\mathbb{R}^d$. The source-to-target map is given by composition $g_2\circ f_1$. 
    On the right (Chained NFs), we construct two NFs $f_1,f_2$ (along with their inverses $g_1,g_2$) transforming the source to standard Gaussian and the source to target, with the source-to-target map given by $f_2$. }
    \label{fig:par_ot}
    \end{center}
    \vskip -0.2in
\end{figure}

To learn the map in Equation~\ref{eq:param_OT_opt}, we first propose a suitable hypothesis set $\mathcal{H}(T)$ as follows. 
We use compositions of NFs, which have recently proven useful in other inference problems~\citep{whang2021composing} to parameterize the function $T$ in~\cref{eq:param_OT_opt}. We show two different ways to construct such a map $T$ to transport from source to target in Figure~\ref{fig:par_ot}, where we build two NFs that are composed in different ways from the base space: triangular NFs (left) and chained NFs (right). 
For Triangular NFs, two NFs $(f_1,f_2)$ describe transformations on $\mathbb{R}^d$ transforming each of the source and target distributions to a standard Gaussian distribution. The corresponding inverse flows $(g_1,g_2)$ describe the inverse transformation of the standard Gaussian on $\mathbb{R}^d$ to the source and target distributions. These can be composed to yield the pushforward map $T^{[\theta_{\text{NF}}]} = g_2\circ f_1$ from source to target, where $\theta_{\text{NF}}$ are learnable parameters. Similarly, for Chained NFs, we learn NFs $(g_1,f_2)$ that transform the standard Gaussian on $\mathbb{R}^d$ to the source distribution and the source distribution to the target distribution, along with the corresponding inverses $(f_1,g_2)$. Here, the pushforward map is $T^{[\theta_{\text{NF}}]} = f$.

Given two datasets $\mathcal{D}_s = \lbrace x_i^{s}\rbrace_{i = 1}^{N_s}$ and $\mathcal{D}_t = \lbrace x_i^{t} \rbrace^{N_t}$ where $x_i^s \sim \mu_s \in \Omega^s$ and $ x_i^{t}\sim \mu_t\in \Omega_t$, respectively, the parameters $\theta_{\text{NF}}$ of the NFs (triangular or chained) should maximize the likelihood 
while also satisfying Equation~\ref{eq:param_OT_opt}. For these two models, we give details on computation of different tasks in Table \ref{tab:jointNFtasks}. Specifically, we outline the procedure for sampling new source or target data points or evaluating the likelihood of new source or target data points via the base density, and for evaluating the pushforward map on new data points.

\begin{table}[t]
\caption{Computation tasks in parOT Triangle and Chained NFs.}
\label{tab:jointNFtasks}
\vskip 0.15in
\begin{center}
\begin{small}
\begin{sc}
\begin{tabular}{ll}
\textbf{Triangle NF} & \\
Task & Procedure  \\
\midrule
Sample $\Omega_s$ & Sample $Z$ and apply $g_1$ \\
Sample $\Omega_t$ & Sample $Z$ and apply $g_2$ \\
Likelihood $x\in \Omega_s$ & $\rho_{Z}(f_1(x))\lvert \det Df_1(x)\rvert$ \\
Likelihood $x\in \Omega_t$ & $\rho_{Z}(f_2(x))\lvert \det Df_2(x)\rvert$ \\
Pushforward Map & $g_2\circ f_1$ \\ 
\midrule
\textbf{Chained NF} & \\
Task & Procedure  \\ \midrule
Sample $\Omega_s$ & Sample $Z$ and apply $g_1$ \\
Sample $\Omega_t$ & Sample $Z$ and apply $f_2\circ g_1$ \\
Likelihood $x\in \Omega_s$ & $\rho_{Z}(f_1(x))\lvert \det Df_1(x)\rvert$ \\
Likelihood $x\in \Omega_t$ & $\rho_{Z}(f_1(g_2(x)))\lvert \det D(f_1\circ g_2)(x)\rvert$ \\
Pushforward Map & $f_2$ \\
\bottomrule
\end{tabular}
\end{sc}
\end{small}
\end{center}
\vskip -0.1in
\end{table}

We can now describe our learning algorithm to solve Equation~\ref{eq:param_OT_opt} as
\begin{equation}
    \widehat{\theta_{\text{NF}}} = \underset{\theta_{\text{NF}}}{\argmin}\quad\mathcal{L}(\theta_{\text{NF}};\mathcal{D}_s^{\text{train}}, \mathcal{D}_t^{\text{train}}, \mathcal{D}_p^{\text{train}}),
\end{equation}
where the loss function $\mathcal{L}$ is the sum of multiple terms,
\begin{equation}
\begin{split}
    \mathcal{L} &= \lambda_{\text{Paired}}\, \text{Paired Loss} +  \text{NLL}_{s} +  \text{NLL}_{t} \\ 
        &\hspace{1em}+ \lambda_{\text{IPM}}\, \text{IPM} + \lambda_{\text{Id}}\, \text{Id. Regularization}. \label{eq:loss}
\end{split}
\end{equation}
In Equation~\ref{eq:loss}, $\text{Paired Loss}$ is the norm discrepancy of the pushforward evaluated on the paired data set, $\text{NLL}_s,\text{NLL}_t$ are the source and target NLLs from the NFs, $\text{IPM}$ is the integral probability metric between the estimated pushforward and the target distribution, and $\text{Id. Regularization}$ is a regularization term for the NF map, similar in spirit to the cost function in OT.
We define these terms explicitly in terms of computable quantities in~\cref{thm:param_OT_opt_wass_data}.
\begin{theorem}\label{thm:param_OT_opt_wass_data}
Given data sets $\mathcal{D}_s,\mathcal{D}_t$ and $\mathcal{D}_p$ and the hypothesis set $\mathcal{H}(T)$ describing the triangular NF as above the solution to the learning problem~\cref{eq:param_OT_opt}, if it exists, on the triple  $(\mathcal{D}_s,\mathcal{D}_t,\mathcal{D}_p)$ is given by by minimizing equation~\ref{eq:loss} where,

{\small
\begin{equation*}
    \text{Paired Loss}(\theta_{\text{NF}},\mathcal{D}_p) = \mathbb{E}_{(x^s,x^t)\sim\mathcal{D}_p} (\lvert g_2(f_1(x^s)) - x^{t}\rvert)^{p}
\end{equation*}
\begin{equation*}
    \text{NLL}_{s}(\theta_{\text{NF},s},\mathcal{D}_s) = -\mathbb{E}_{x\sim \mathcal{D}_s}\left[\log\left(\rho_{Z}(f_1(x_i^s))\lvert\det Df_1(x_i^s)\rvert\right)\right]
\end{equation*}
\begin{equation*}
    \text{NLL}_{t}(\theta_{\text{NF},t},\mathcal{D}_t) = -\mathbb{E}_{x\sim\mathcal{D}_t}\left[\log\left(\rho_{Z}(f_2(x_i^t))\lvert\det Df_2(x_i^t)\rvert\right)\right].
\end{equation*}
}
{\small
\begin{equation*}\label{eq:IPM_NF}
    \begin{split}
    \text{IPM}(\theta_{\text{NF}},\mathcal{D}_s,\mathcal{D}_t) &= \lvert \mathbb{E}_{x\sim\mathcal{D}_t}\left[\phi(x)\right] \\
  & \quad - \mathbb{E}_{x\sim\mathcal{D}_s}\left[\phi(g_2(f_1(x)))\right]\rvert.\\
  \end{split}
\end{equation*}}

{\small
\begin{equation*}
    \text{Id. Reg}(\theta_{\text{NF}},\mathcal{D}_s) = \mathbb{E}_{x\sim\mathcal{D}_s}\left[(g_2(f_1(x)) - x)^{2}\right]
\end{equation*}
}

\end{theorem}
See~\cref{sec:app_learning_algo} in the Appendix for details and proof. In practice, the size of the paired dataset $N_p$ will be much smaller than size of source and target datasets $N_s,N_t$ (e.g., by a factor of 5-10). 


\subsection{Domain Translation and Adaptation}
Domain translation is the process of transforming one data domain to more closely resemble another, and it has many potential downstream applications.
One of the potential applications is domain adaptation; here, we assume that labeled source domain data is available for classification or regression, but labeled target domain data is not available.
The goal is to achieve a predictive model that is accurate on the target domain despite the lack of labels.
As in many domain adaptation works, we assume covariate shift~\citep{kouw2018introduction}; that is, $p(Y | T(X_s)) = p(Y | X_t)$.
To achieve domain adaptation, we first transform our labeled source data to the target domain, then train a supervised learning method, then apply it to the target domain data.
For instance, in the Chained NF, from source data $\mathcal{D}_s = \{x_i^s, y_i^s\}^{N_s}$, we create training data $\tilde{\mathcal{D}}_s = \{f_2(x_i^s), y_i^s\}^{N_s}$ for the supervised learning algorithm.
The goal is for $\tilde{\mathcal{D}}_s$ to represent the labeled target data (though we cannot observe the labels).

\section{Results}
In this section, we present experiments and results on three datasets: two benchmark data sets (a simulated two-dimensional Gaussian mixture and the Two Moons dataset, Section~\ref{sec:results_benchmark}) and a real-world scientific application (ChemCam, Section~\ref{sec:e2m_dataset}).
Our results represent example applications of parOT; specifically, we demonstrate successful transformation (\cref{sec:intro}, Task 1) and probabilistic modeling (\cref{sec:intro}, Task 3) while illustrating the utility of map constraints (\cref{sec:intro}, Task 2). For the synthetic datasets, we compare our results to the methods identified in Table \ref{tab:benchmark_methods}.
In the ChemCam application, we compare parOT to a domain-knowledge transformation used by domain experts.


\subsection{Benchmark experiments} \label{sec:results_benchmark}
\begin{figure}[t]
    \vskip 0.2in
    \begin{center}
    \includegraphics[width=0.6\linewidth]{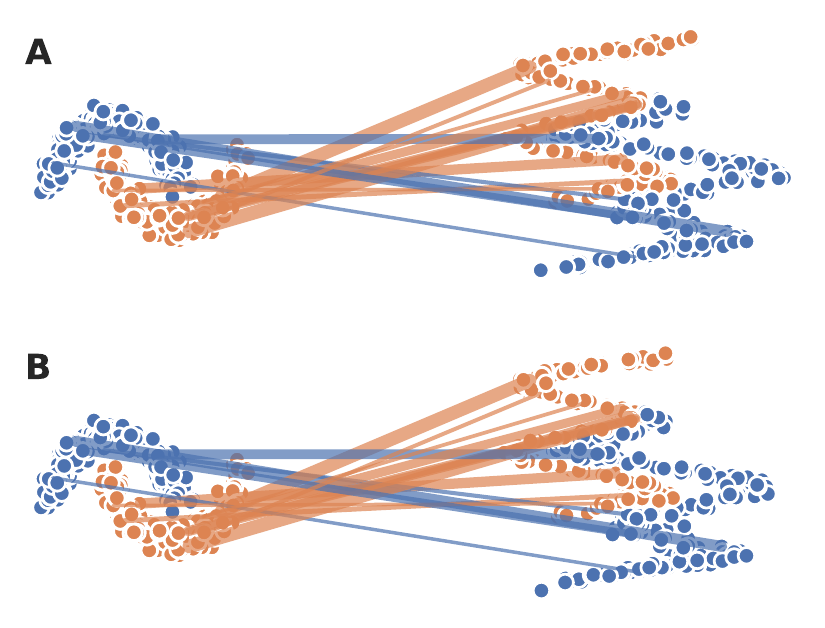}
    \caption{A. Two Moons data (left) with true nonlinear transform (right), colored by class label. B. Two Moons data (left) with parOT estimated nonlinear transform (right). Lines show movement of individual points, indicating agreement between the maps.} 
    \label{fig:two_moons_trans}
    \end{center}
    \vskip -0.2in
\end{figure}

To compare parOT to related methods and to evaluate the effect of hyperparameters, we consider two two-dimensional synthetic data sets: a two-component Gaussian mixture distribution and the Two Moons dataset~\citep{twomoons}.
In both cases, the source data is subjected two two different ground truth transformation maps $F$ (linear and nonlinear; Table~\ref{tab:true_maps}) to create target domain data.
An example of $F_{\text{nonlinear}}$ applied to data sampled from the Two Moons dataset is shown in~\cref{fig:two_moons_trans}.A.
For the mixture of Gaussians dataset, we sample 1,000 training data points from a two-component Gaussian mixture on $\Omega_s$ with means $[-2, 0]$ and $[2, 0]$, marginal variances $[1.0, 1.0]$ and $[0.9, 0.9]$, and correlations $0.7$ and $-0.24$ and use 100 points in the test set.
For the Two Moons dataset, we sample 2,000 training data points from a noisy standard Two Moons shape (with Gaussian noise $\sigma=0.1$) on $\Omega_s$ and use 500 points in the test set.
For each experiment, a proportion of the training data is treated as paired, while the rest is not; this means a small portion of the data satisfies $F(x^s_k) = x^t_k$.
For the methods that can utilize paired training data, we evaluate the methods both with and without paired training data.
For our method, we selected hyperparameters yielding the best performance from a grid search (see Appendix~\ref{sec:app_nf_hyper}, \ref{sec:app_ot_hyper} for more details).

\begin{table}[]
    \centering
    \caption{True linear and nonlinear maps used to generate target data from source data in the benchmark data sets.}
    \label{tab:true_maps}
    \begin{tabular}{l|l|l}
        Dataset & $F_{\text{linear}}$ & $F_{\text{nonlinear}}$ \\ \midrule
        Mixture of Gaussian & Rotate $\pi/4$, scale $[1.0, 0.7]$, shift $[2.0, -0.5]$ & $f(x,y) = (\sin(2 \pi x) + \exp(y), x)$ \\
        Two Moons & Rotate $\pi/4$ & $f(x,y) = (\sin(2 \pi x) + \exp(y), x)$
    \end{tabular}
\end{table}

\subsubsection{Comparison to related methods}
To compare our method to related methods, we calculate pairwise accuracy of source-to-target mapped points and the corresponding target point.
That is, the test data is paired, meaning a test item $(x^s_k, x^t_k)$ satisfies $x^t_k = F(x^s_k)$ where $F$ is the ground truth mapping function.
We estimate the error between an estimated out-of-sample map $\hat{T}$ and the true map $F$ as
\begin{align}
    \rho_\text{MSE}(\hat{T}) = \frac{1}{N} \sum_{k=1}^N \left( \hat{T}(x^s_k) - x^t_k \right)^2. \label{eq:mse}
\end{align}

\begin{table}
    \centering
    \caption{Map accuracy on a paired test set, measured by MSE as defined in~\cref{eq:mse}, for the Mixture of Gaussian benchmark under two true source to target maps (linear and nonlinear). For each method, we indicate whether paired data was used during training.}
    \label{tab:mse_map}
    \begin{subtable}
        \centering
        \begin{tabular}{l|c|r|r} 
            Method & Paired data? & $F_\text{Linear}$ & $F_\text{Nonlinear}$ \\ \hline
            parOT & Y & 0.0009 & 0.25 \\ 
            Linear KGOT & Y & 0.01 & 0.48 \\ 
            Kernel KGOT & Y & 0.09 & 3.40 \\ 
            parOT & & 0.34 & 5.40 \\ 
            Linear OT & & 0.52 & 5.01 \\ 
            Kernel OT & & 1.14 & 5.43 \\ 
            1NN & & 0.40 & 6.62 \\ 
            SWOT-Flow & & 2.14 & 8.47 \\ 
            Rout & & 0.81 & 8.43 \\ 
        \end{tabular}
    \end{subtable}
    \begin{subtable}
        \centering
        \caption{Similar to~\cref{tab:mse_map}, but for the Two Moons benchmark.}
        \label{tab:mse_map2}
        \begin{tabular}{l|c|r|r} 
            Method & Paired data? & $F_\text{Linear}$ & $F_\text{Nonlinear}$ \\ \hline
            parOT & Y & 0.06 & 0.01 \\ 
            Linear KGOT & Y & 0.02 & 0.33 \\ 
            Kernel KGOT & Y & 0.03 & 0.24 \\ 
            parOT & & 0.37 & 0.27 \\ 
            Linear OT & & 0.31 & 2.39 \\ 
            Kernel OT & & 1.13 & 1.81 \\ 
            1NN & & 0.42 & 2.62 \\ 
            SWOT-Flow & & 0.65 & 2.31 \\ 
            Rout & & 0.38 & 2.71 \\ 
        \end{tabular}
    \end{subtable}
\end{table}

Table~\ref{tab:mse_map} gives the results in terms of $\rho_\text{MSE}$ (\cref{eq:mse}) for the two benchmark data sets. 
Generally, methods that utilize paired data perform better in recovering the true map than methods that do not.
When utilizing paired training data, our method outperforms the other methods in all but one case (Mixture of Gaussians with true map $F_\text{linear}$). 
Without paired training data, our method is competitive with and clearly outperforms some of the other methods.
Figure~\ref{fig:two_moons_trans}.B shows the learned transformation corresponding to the ground truth nonlinear transform; our map matches the ground truth transformation very closely.
These results indicate that our method can recover the true map competitively with related methods, but gains in accuracy if paired data is included.

\subsubsection{Probabilistic modeling}
To evaluate the impact of map regularization on the resulting probabilistic model in our framework, we evaluate how well parOT recovers the true pointwise likelihood of each target test set data point for the mixture of Gaussians data. 
Here, we compare relative MSE of the pointwise likelihood under different modeling choices within our framework. 
When no paired data is used for training, our Chained and Triangle NFs achieve 0.018 and 0.042 relative MSE; when 0.2 paired data is used, these improve to 0.009 and 0.006, respectively.
These results suggest that inclusion of paired data improves likelihood estimation in our framework.




\subsubsection{Domain adaptation}
We further demonstrate the utility of the learned transformation for domain adaptation. 
Table \ref{tab:sim_da} shows target test set classification accuracy with and without domain adaptation when an SVM classifier is trained on transformed source data with no paired data (paired proportion 0.0) and with 20\% of the training set consisting of paired data (paired proportion 0.2). 
In both scenarios, parOT achieves higher predictive accuracy than OT or KGOT, greatly surpassing the baseline case (no domain adaptation).

\begin{table}
    \caption{Target test set classification accuracy following domain adaptation by different methods: no domain adaptation (No DA), linear OT or KGOT (L. OT/KGOT), kernel OT or KGOT (K. OT/KGOT), or parOT for Two Moons data under two proportions of paired training data (0.0, 0.2) and a nonlinear transform.}
    \label{tab:sim_da}
    \vskip 0.15in
    \centering
    \begin{adjustbox}{max width=\linewidth}
    \begin{tabular}{l|r|r|}
        Method & Paired prop. 0.0 & Paired prop. 0.2 \\ \toprule
        No DA & 0.281 & -- \\
         L. OT/KGOT & 0.204 & 0.829  \\
     K. OT/KGOT & 0.441  & 0.816 \\
         parOT &\textbf{0.756} & \textbf{0.939} \\ \bottomrule
    \end{tabular}
    \end{adjustbox}
    \vskip -0.1in
\end{table}

\subsection{Earth-to-Mars} \label{sec:e2m_dataset}
To evaluate parOT on the scientific application of interest, we use a source data set of 2,442 calibration spectra collected on Earth corresponding to 489 unique materials.
Each spectrum consists of 5,205 intensity values across a range of wavelengths.
For each spectrum, the true composition of the material is given in terms of weight percent values of eight oxides that are used to describe the composition of geological samples. 
The unlabeled target data set consists of 23,649 spectra collected on Mars. In addition, we have 309 rover calibration spectra in which the spectra are collected on Mars, but have known compositions, corresponding to six unique materials (Norite, Picrite, Shergottite, NAu-2-low-s, NAu-2-mid-s, and NAu-2-high-s).
Because the spectra are high-dimensional, we learn the map in a lower-dimensional latent space via an encoder/decoder.
Additional details about the data and modeling details are given in Appendix sections~\ref{sec:app_e2m_data} and~\ref{sec:app_e2m_nf}.
We evaluate parOT for Earth-to-Mars transformation quality (Section~\ref{sec:e2m_trans}) and domain adaptation (Section~\ref{sec:e2m_da}); in both cases, we compare to a domain-knowledge (DK) Earth-to-Mars transformation proposed in~\citealt{clegg2017recalibration}.
Unlike our framework, the DK transformation is not algorithmic and data-driven; instead, it is derived by domain experts with the use of a limited number of calibration target spectra.
Finally, we demonstrate a potential use case of the probabilistic model in Section~\ref{sec:e2m_ol}.

\begin{table}
    \caption{RMSE of averaged transformed Earth-to-Mars spectra (relative to Mars measurements); parOT outperforms the standard DK transform by a factor of two.}
    \label{tab:e2m_trans}
    \vskip 0.15in
    \centering
    \begin{tabular}{l|r|r}
        Target & parOT & DK trans. \\ \toprule
        Norite & \textbf{0.007} & 0.012 \\
        Picrite & \textbf{0.005} & 0.010 \\
        Shergottite & \textbf{0.004} & 0.011 \\
        NAu-2-low-s & \textbf{0.005} & 0.010 \\
        NAu-2-mid-s & \textbf{0.005} & 0.011 \\
        NAu-2-high-s & \textbf{0.006} & 0.012
    \end{tabular}
    \vskip -0.1in
\end{table}

\subsubsection{Earth-to-Mars Transformation}\label{sec:e2m_trans}
\begin{figure}[t]
    \vskip 0.2in
    \begin{center}
    \includegraphics[width=0.5\linewidth]{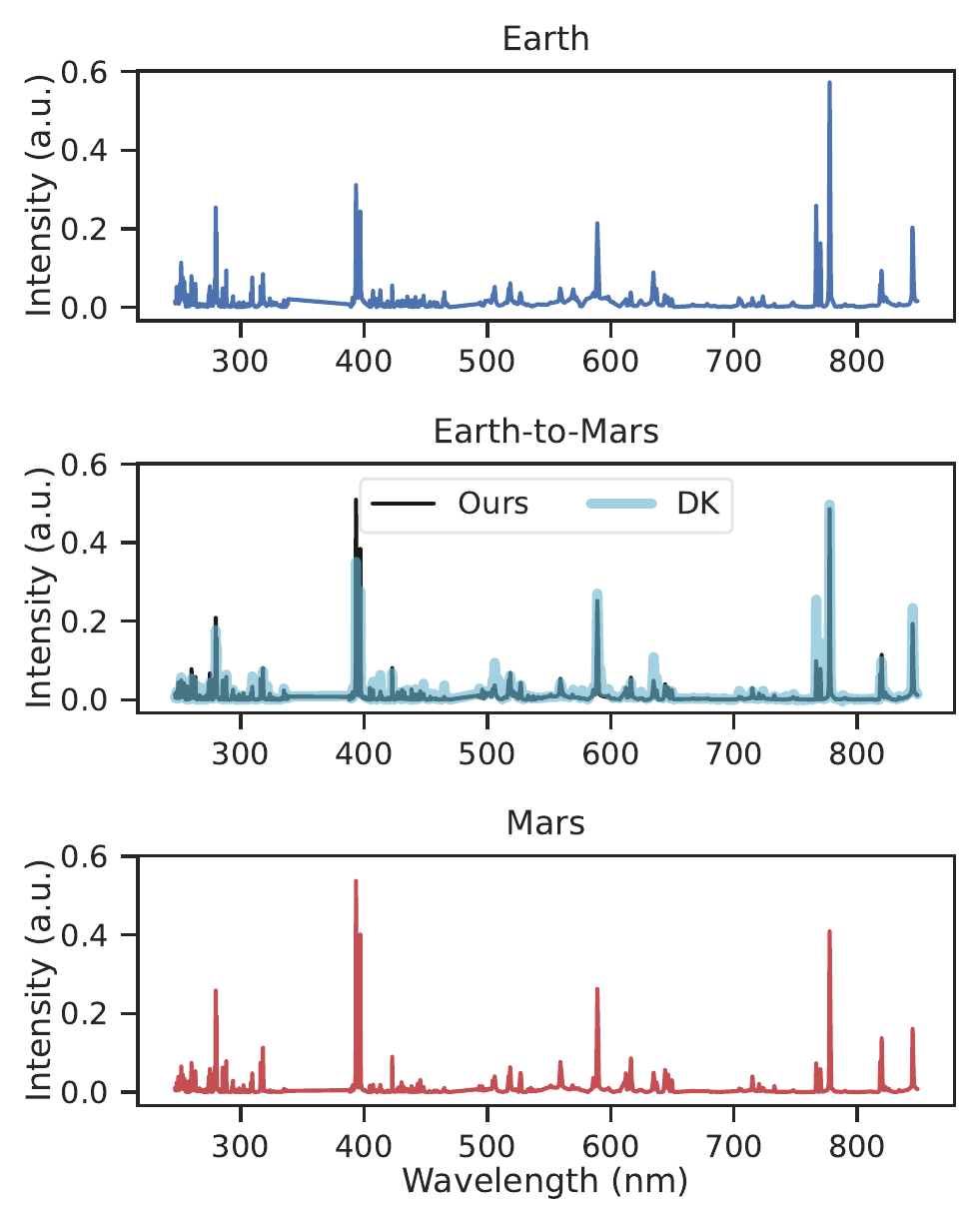}
    \caption{Shergottite spectra measured on Earth (top) and Mars (bottom), with the middle row showing the Earth-to-Mars transform by parOT (black) and the DK method (cyan). The DK method is notably deficient in capturing differences between Earth and Mars at some wavelengths (e.g., near 393nm and 766nm).}
    \label{fig:e2m_spec}
    \end{center}
    \vskip -0.2in
\end{figure}

\begin{figure}
    \vskip 0.2in
    \begin{center}
    \includegraphics[width=0.5\linewidth]{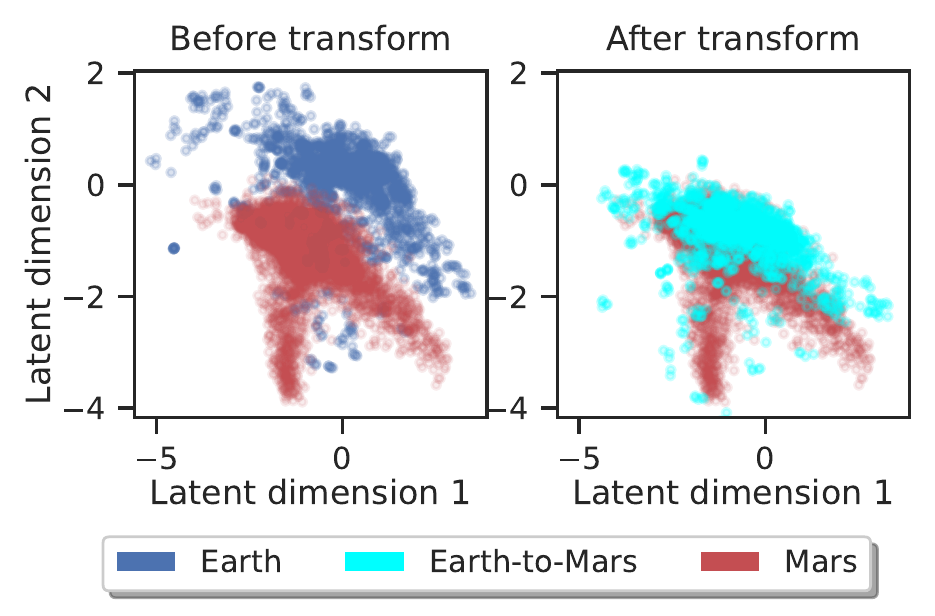}
    \caption{Two-dimensional representation of Earth and Mars data before parOT transform (left) and after parOT transform (right).}
    \label{fig:e2m_trans_2d}
    \end{center}
    \vskip -0.2in
\end{figure}

Here, we evaluate whether transformed Earth calibration spectra match the corresponding Mars rover calibration spectra.
For each of the six targets, we first transform the Earth spectra to the Mars domain; because we have different numbers of measurements on Earth and Mars, we take the averaged Earth-to-Mars spectra for each target and compare to the averaged Mars spectra via root-mean squared error (RMSE).
Table~\ref{tab:e2m_trans} shows that parOT achieves lower RMSE (by approximately a factor of two) compared to the DK transformation. 
Figure~\ref{fig:e2m_spec} shows the mean Shergottite spectrum on Earth and Mars compared to the mean transformed spectrum.
There are notable deficiencies in the DK transform near 393nm and 766nm, as the predicted Earth-to-Mars spectrum (cyan) has much lower or much higher peaks compared to the true Mars spectrum; these lines correspond to concentrations of Ca and K.
In contrast, parOT (black) more closely matches the Mars spectrum.
We observed similar results for the other five calibration targets.
Figure~\ref{fig:e2m_trans_2d} shows the effect of our learned Earth-to-Mars transformation represented in in a two-dimensional latent space (left: before transform, right: after transform). Before transform, there is a clear distribution shift between Earth and Mars data, while the distributions appear to be more well-aligned after transformation. We note that due to paired information constraints and regularization, we don't expect complete alignment of the distributions, but observe closer alignment of the empirical mass.

\begin{table*}[t]   
    \caption{Mars calibration spectrum MSE (SD) of standardized composition value by oxide (columns) for two domain adaptation methods compared to a model with no domain adaptation (top row). parOT generally outperforms the DK method and with smaller variation (bold indicates the best MSE among DK and parOT, with the top row for reference only).}
    \label{tab:e2m_da}
    \vskip 0.15in
    \centering
    \begin{adjustbox}{max width=\linewidth}
    \begin{tabular}{l|rrrrrrrr}
         & Al$_2$O$_3$ & CaO & FeOT & K$_2$O & MgO & Na$_2$O & SiO$_2$ & TiO$_2$ \\ \toprule
        No trans. & 0.387 (0.481) & 0.689 (0.503) & 0.621 (1.114) & 0.017 (0.026) & 0.214 (0.226) & 0.058 (0.056) & 0.609 (0.685) & 0.086 (0.066)\\ \arrayrulecolor{gray}\hline 
        DK trans. & 0.402 (0.425) & 0.398 (0.348) & 2.055 (1.872) &  \textbf{0.004} (0.006) & 0.261 (0.305) &  \textbf{0.010} (0.013) & 1.014 (0.868) & 0.145 (0.085)\\  
        parOT & \textbf{0.196} (0.167) & \textbf{0.199} (0.225) &  \textbf{0.962} (1.074) & 0.043 (0.084) &  \textbf{0.184} (0.222) & 0.018 (0.024) &  \textbf{0.758} (0.753) &  \textbf{0.006} (0.016)\\ 
    \end{tabular}
    \end{adjustbox}
    \vskip -0.1in
\end{table*}

\subsubsection{Earth-to-Mars Domain Adaptation}\label{sec:e2m_da}
To evaluate how parOT affects domain adaptation, we train neural networks to predict chemical composition (in terms of eight oxide weight percent values for oxides Al$_2$O$_3$, CaO, FeOT, K$_2$O, MgO, Na$_2$O, SiO$_2$, and TiO$_2$).
The networks are trained on labeled Earth spectra with no transformation, or with labeled Earth spectra transformed with parOT or the DK transform.
After training, we evaluate the models on the six calibration targets from the Mars rover (for which composition values are known) and summarize across targets by the MSE of the oxide weight percent; for ease of comparison, the oxide weight percent values are standardized by the training mean and standard deviation as the typical amount of each oxide spans a large range.
In addition to MSE, we compute the standard deviation (SD) of the per-observation MSE within each oxide group to measure the variability of each model on the full set of Mars calibration data.
Table~\ref{tab:e2m_da} shows that in terms of MSE, parOT outperforms the DK method on six of the eight oxides, and generally has lower SD, indicating lower variance in predictive accuracy across observations.
For the remaining two oxides (K$_2$O and Na$_2$O), the models based on the DK transform perform better than those based on our transform; yet in the case of Na$_2$O, parOT still outperforms no transformation.
These results suggest that parOT can provide useful domain adaptation, but research into optimal regression models for this task is ongoing.

\subsubsection{Earth-to-Mars Probabilistic Model}\label{sec:e2m_ol}
\begin{figure}
    \vskip 0.2in
    \centering
    \includegraphics[width=0.5\linewidth]{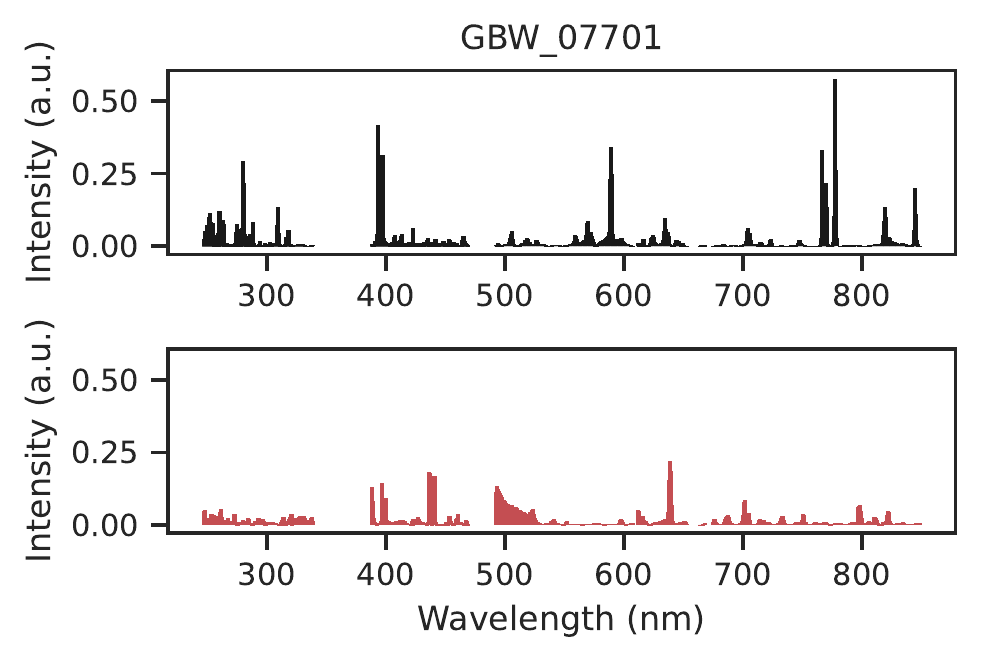}
    \caption{Example spectra for target GBW\_07701; the top row is considered in-distribution (according to the likelihood within the learned probabilistic model), while the bottom row is considered an outlier (negative log likelihood in the top 0.99 quantile). Clearly, the bottom spectrum deviates from the expected spectral intensity, perhaps due to problems during data collection or processing.}
    \label{fig:e2m_ol}
    \vskip -0.2in
\end{figure}

Here, we demonstrate the use of the parOT probabilistic model for outlier detection.
Based on the induced density of the Earth samples (from the simple base density under the NF transform), we evaluate pointwise NLL for each spectra.
Points with unusually high NLL have low likelihood under the learned density and may be considered outliers.
Figure~\ref{fig:e2m_ol} shows a representative example with NLL greater than the 0.99 quantile in the Earth dataset (target GBW\_07701, a synthetic silicate); the top row shows spectra from this target not considered outliers in the probabilistic model, while the bottom row shows the outlier spectrum.

\section{Conclusions}
In this work, we introduce parOT, a novel semi-supervised learning framework for pushforwards via parametric OT.
Our framework, based on composed NFs, has appealing mathematical properties compared to existing methods; in particular, it directly addresses three key tasks: transformation, map constraints, and probabilistic modeling.
We demonstrate in three data sets (including a high-dimensional real-world scientific application) that our framework outperforms existing OT methods on a variety of tasks.
Our framework offers a new approach to pushforward estimation that could be useful in domain translation and adaptation as well as generative modeling.
The current framework relies on a shared low-dimensional space; in future work, we will extend our framework to data with many dimensions, or unequal dimensions (similar to~\citealt{rout2021generative}).
In addition, inductive biases of different NF models could affect the results; further study will elucidate which types of NFs are most appropriate for the three tasks presented.
We could also consider other parameterizations entirely, including diffusion models.
Similarly, inclusion of more sophisticated domain-knowledge constraints (such as physics-based constraints) is an area for future exploration.
While our approach solves task 3 (probabilistic model), noted deficiencies of NFs for out-of-distribution detection~\citep{kirichenko2020normalizing} suggest more thorough study is needed.
Finally, in this work, we used a two-step approach to domain adaptation (transformation followed by training), but we can include the target supervised learning model in the learning procedure to leverage feature spaces better suited for downstream prediction.


\bibliography{refs}
\bibliographystyle{icml2023}

\clearpage  
\begin{appendices}
\section{Learning Algorithm}\label{sec:app_learning_algo}

 We  describe in full detail the learning algorithm for~\cref{eq:param_OT_opt} for the particular case when $\gamma_{\mathcal{F}} = W^1$ and define all the terms in~\cref{eq:loss}. Additionally, we will describe the complete framework for~\cref{eq:param_OT_opt} when the source and target dimensions differ. Let $\mathbb{P},\mathbb{Q}$ be two probability measures defined on the same sigma algebra over a set $M$, the integral probability metric $\gamma_{\mathcal{F}}(\mathbb{P},\mathbb{Q})$ is given by
\begin{equation}\label{eq:IPM}
    \gamma_{\mathcal{F}}(\mathbb{P},\mathbb{Q}) = \underset{\phi\in\mathcal{F}}{\sup}{\left\lvert \int_{M}\phi d\mathbb{P} - \int_M\phi d\mathbb{Q}\right\rvert}
\end{equation}
where $\mathcal{F}$ is a suitable class of real valued measurable functions on $M$. When  $\mathcal{F}$ is the class of $1-$Lipschitz functions on $M$, denoted by $\text{Lip}^{1}(M)$ i.e.~the set $\lbrace \phi: \lvert\lvert \phi \rvert\rvert_{L} \leq 1\rbrace$ we get the Kantorovich metric~\citep{sriperumbudur2009integral}, which by the Kantorovich-Rubenstein duality is equal to the $1-$Wasserstein distance between $\mathbb{P}$ and $\mathbb{Q}$~\cite{arjovsky2017wasserstein}. 

We now prove~\cref{thm:param_OT_opt_wass_data}.
\begin{proof}
    When $\gamma_{\mathcal{F}} = W^1$, by taking $\mathbb{Q} = T\sharp\mu_{s}$ and $\mathbb{P} = \mu_t$ and $M = \Omega_t$ in~\cref{eq:param_OT_opt} we get
    \begin{equation*}
    \begin{split}
            T_0 &= \underset{T:\Omega_s\to \Omega_t}{\argmin} H(T) \\
            &+ \lambda\underset{\begin{subarray}{c}
  \phi:\Omega_t\to \mathbb{R} \\
  \phi\in\text{Lip}^{1}(\Omega_t)
  \end{subarray}}{\sup} \lvert \mathbb{E}_{x\sim\mu_t}[\phi(x)] - \left.\mathbb{E}_{x\sim {T}\sharp\mu_s}[\phi(x)]  \right) \rvert.
    \end{split}
\end{equation*}
    Evaluating $(\lvert\lvert T - F \rvert\rvert_{p})^{p}$ on the (source projection of the paired) dataset $\mathcal{D} = \lbrace x_k\in \Omega_s : x_k\in \mathcal{D}_p\rbrace$ we get,
    $\mathbb{E}_{x\sim \mathcal{D}}\lvert T(x) - F(x)\rvert^p$ which gives us \[\frac{1}{N_p}\sum_{k = 1}^{N_p}(\lvert T(x_k^s) - F(x_k^s)\rvert)^{p}.\] Using the fact that $F(x_k) = x_k^t$ for $x_k\in \mathcal{D}_p$ we get the $\text{Paired}$ Loss. Evaluating $(\lvert\lvert T - I \rvert\rvert_{p})^{p}$ on $\mathcal{D}_s$ we get the $\text{Id. Reg}$ Loss.
    
    For the pushforward constraint, we use the following fact about integration of function w.r.t pushforwards~\citep{bogachev2007measure}.
\begin{align*}
    \int_{A\subset \Omega_t}\phi d\, T\sharp\mu_s = \int_{T^{-1}(A)\subset \Omega_s}\phi\circ T\,d\mu_s
\end{align*} to write
\begin{align*}
    \mathbb{E}_{x\sim T\sharp\mu_s}[\phi(x)] = \mathbb{E}_{x\sim\mu_s}[\phi(T(x))].
\end{align*}
Evaluating $\mathbb{E}_{x\sim \mu_t}[\phi(x)] $ and $\mathbb{E}_{x\sim \mu_s}[\phi(T(x))]$ on $\mathcal{D}_t$ and $\mathcal{D}_s$ respectively we get the $\text{IPM Loss}$.
\end{proof}

For the case when the dimensions of the source and target spaces are equal to $d$, under the assumption that the target distribution $\mu_t$ is equal to the pushforward of the source distribution $\mu_s$ under a diffeomorphic map $F$ we can model $T$ as normalizing flow maps as in~\cref{fig:par_ot}. For e.g.~by taking $T$ to be the Triangular NF $T = g_2\circ f_1$ with corresponding parameters $\theta_{\text{NF}} = \lbrace\theta_{\text{NF},s}$ and $\theta_{\text{NF},t}\rbrace$ for the normalizing flows $(f_1,g_1)$ and $(f_2,g_2)$ respectively we have the following terms in the Loss function $\mathcal{L}$ in~\cref{eq:loss}.
{\small
\begin{equation}
    \text{Paired Loss}(\theta_{\text{NF}},\mathcal{D}_p) = \frac{1}{N_p}\sum_{k = 1}^{N_p}(\lvert g_2(f_1(x_k^s)) - x_k^{t}\rvert)^{p}
\end{equation}
\begin{equation}
    \text{NLL}_{s}(\theta_{\text{NF},s},\mathcal{D}_s) = -\mathbb{E}_{x\sim \mathcal{D}_s}\left[\log\left(\rho_{Z}(f_1(x_i^s))\lvert\det Df_1(x_i^s)\rvert\right)\right]
\end{equation}
\begin{equation}
    \text{NLL}_{t}(\theta_{\text{NF},t},\mathcal{D}_t) = -\mathbb{E}_{x\sim\mathcal{D}_t}\left[\log\left(\rho_{Z}(f_2(x_i^t))\lvert\det Df_2(x_i^t)\rvert\right)\right].
\end{equation}
}
{\small
\begin{equation}\label{eq:IPM_NF}
    \begin{split}
    \text{IPM}(\theta_{\text{NF}},\mathcal{D}_s,\mathcal{D}_t) &= \lvert \mathbb{E}_{x\sim\mathcal{D}_t}\left[\phi(x)\right] \\
  & \quad - \mathbb{E}_{x\sim\mathcal{D}_s}\left[\phi(g_2(f_1(x)))\right]\rvert.\\
  \end{split}
\end{equation}}

{\small
\begin{equation}
    \text{Id. Reg}(\theta_{\text{NF}},\mathcal{D}_s) = \mathbb{E}_{x\sim\mathcal{D}_s}\left[(g_2(f_1(x)) - x)^{2}\right]
\end{equation}
}
In order to use the loss function~\cref{eq:loss} into an algorithm we need to parameterize the space of $1-$ Lipschitz function (for e.g.~by a neural network) and satisfy the Lipschitz norm constraint. We can do this as in~\cite{arjovsky2017wasserstein} by using a 1-hidden layer neural network mapping inputs in $\mathbb{R}^d$ to values in $\mathbb{R}$ and enforcing its gradient to have a $2-$norm of $1$. Thus, if $\theta_{\phi}$ are the parameters of $\phi$, we can re-write the IPM loss in~\cref{eq:IPM_NF} as follows
{\small
\begin{equation}\label{eq:IPM_NF_Constraint}
    \begin{split}
    \text{IPM}(\theta_{\text{NF}},\theta_{\phi}, \mathcal{D}_s,\mathcal{D}_t) &= 
    \lvert \mathbb{E}_{x\sim\mathcal{D}_t}\left[\phi(x)\right] \\
  & \quad - \mathbb{E}_{x\sim\mathcal{D}_s}\left[\phi(g_2(f_1(x)))\right] \rvert\\
  & +\lambda \mathbb{E}_{x\sim\mathcal{D}_s}\left(\lvert\lvert \nabla_{\Omega_t}(\phi\circ \left. g_2\circ f_1)\rvert\rvert^2 - 1\right)\right)\\
  \end{split}
\end{equation}}
This gives us the following $\min-\max$ algorithm to solve~\cref{eq:param_OT_opt}
{\small
\begin{equation}
    \widehat{\theta_{\text{NF}}} = \underset{\theta_{\text{NF}}}{\argmin}\,\underset{\theta_{\phi}}{\argmax}\quad\mathcal{L}(\theta_{\text{NF}},\theta_{\phi};\mathcal{D}_s^{\text{train}}, \mathcal{D}_t^{\text{train}}, \mathcal{D}_p^{\text{train}}),
\end{equation}}
where the loss function is given by~\cref{eq:loss}. In practice, as we do in this paper, we can use sliced Wasserstein distance~\citep{deshpande2018generative} to compute~\cref{eq:IPM_NF}, avoiding the min-max formulation by directly approximating the Wasserstein distance. 

\begin{figure}
    \centering
    \includegraphics[width=0.5\linewidth]{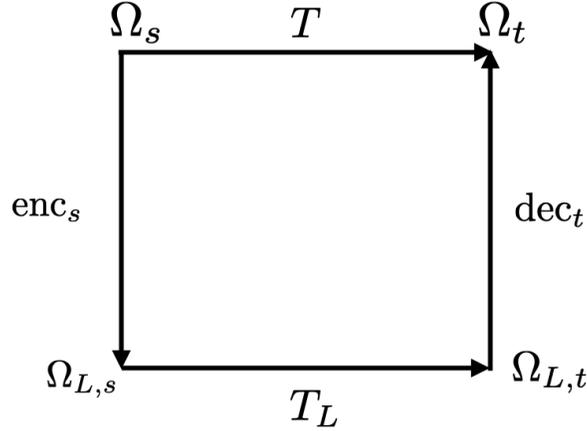}
    \caption{In higher dimensions and/or when $\Omega_s,\Omega_t$ are subsets of different dimensional spaces we construct the pushforward map $T$ by decomposing it into $3$ maps. An encoder-decoder pair $(\text{enc}_s,\text{dec}_s)$ on $\Omega_s$ to the latent space $\Omega_{L,s}$, an encoder-decoder pair $(\text{enc}_t,\text{dec}_t)$ on $\Omega_t$ to the latent space $\Omega_{L,t}$ and a diffeomorphic map from $\Omega_{L,s}$ to $\Omega_{L,t}$ where the source and target latent spaces are subset of $\mathbb{R}^d$. The source-to-target map is given by $T = \text{dec}_t\circ T_L\circ\text{enc}_s$. }
    \label{fig:par_ot_full}
\end{figure}

We now discuss our framework in the \emph{general} case when source and target spaces have different dimensions or are high dimensional. In such a case, we can project the two data spaces on to a latent space of the same dimension and consider the diffeomorphic pushforward between the latent spaces in exactly the same fashion we have done so far. Let $\Omega_s\subset\mathbb{R}^{d_s}$ and $\Omega_t\subset\mathbb{R}^{d_t}$. Consider the latent spaces $\Omega_{L,s},\Omega_{L,t} \subset \mathbb{R}^d$ where $d \leq \min(d_s,d_t)$. Associated with each latent space is a pair of encoder and decoder maps, that is $\text{enc}_s:\Omega_s \to \Omega_{L,s}$ and $\text{enc}_t:\Omega_t\to\Omega_{L,t}$ respectively (with the corresponding decoder maps). If $T$ is any source-to-target map, we can decompose $T$ as the composition $T = \text{dec}_t\circ T_L \circ \text{enc}_s$ where $T_L$ is a diffeomorphism on the latent spaces as shown in~\cref{fig:par_ot_full}. With $T$ given by this decomposition, the optimization framework in~\cref{eq:param_OT_opt} remains exactly the same. The learning algorithm gets modified to include loss terms from the encoder and decoder pair whereas the normalizing flow framework is exactly applied to the function $T_L$. 

Let  $\theta_{\text{enc-dec}}$ be the parameters of the encoder-decoder pairs on $\Omega_s,\Omega_t$ and $\theta_{\text{NF}}$ be the parameters of the normalizing flow map $T_L$. Let $\theta = \lbrace \theta_{\text{NF}}, \theta_{\text{enc-dec}}\rbrace$ be the set of all learnable parameters. Our learning algorithm to solve~\cref{eq:param_OT_opt} in this general situation can now be described as the solution to the following equation 

\begin{equation}
    \widehat{\theta} = \underset{\theta}{\argmin}\quad\mathcal{L}(\theta_{\text{NF}},\theta_{\text{enc-dec}};\mathcal{D}_s^{\text{train}}, \mathcal{D}_t^{\text{train}}, \mathcal{D}_p^{\text{train}}),
\end{equation}
where the loss function $\mathcal{L}$ is the sum of multiple terms,
\begin{equation}
\begin{split}
    \mathcal{L} &= \lambda_1\, \text{Paired Loss} + \lambda_2\, \text{NLL}_{s} + \lambda_3\, \text{NLL}_{t} \\ 
        &\hspace{1em}+ \lambda_4\, \text{IPM} + \lambda_5\, \text{Id. Regularization}\\
        &\hspace{1em}+ \lambda_6\, \text{Recon. Loss}_s + \lambda_7\, \text{Recon. Loss}_t.
        \label{eq:loss_gen}
\end{split}
\end{equation}
Here, $\text{Recon. Loss}_s$ and  $\text{Recon. Loss}_t$ are the standard \emph{reconstruction} losses on $\Omega_s,\Omega_t$ respectively. 

We develop our framework in \texttt{PyTorch}; research code is included as a supplement to the paper, but we will release a clean, user-friendly open-source version upon acceptance. In addition to standard Python packages, we use NF elements from~\url{https://github.com/VincentStimper/normalizing-flows}.

\section{Map Estimation with KGOT}
We modified open-source KGOT code (\url{https://github.com/XJTU-XGU/KPG-RL}) to estimate linear or kernel maps using the block coordinate descent approach of \texttt{PythonOT MappingTransport}.
During the map solution step, we use least squares the same as \texttt{PythonOT}, but imposing a mask on the coupling matrix to enforce identity mapping of paired points.
During the coupling matrix solution step, we use a modified version of \texttt{PythonOT gcg} function for generalized conditional gradient descent.
In addition to masking the coupling matrix, we add the regularization term on the coupling from~\cite{gukeypoint} during the OT solution.
We found that this formulation is less stable than the original \texttt{MappingTransport} optimization problem and can suffer from problems during optimization for some hyperparameter choices; it appears that joint estimation of the map with the pairing constraints is a non-trivial extension of \texttt{MappingTransport}. 
While our results with modified KGOT suggest inclusion of paired information is generally helpful for map estimation, improving this algorithm could be a focus of future work.

\section{Diffeomorphic Map on 2$d$ Details}
We describe details of the data used in Section~\ref{sec:results_benchmark} as well as the algorithms and hyperparameters.

\subsection{NF Hyperparameters} \label{sec:app_nf_hyper}
We use a grid search over the hyperparameters described in Table~\ref{tab:app_mog_hyper} and fix other hyperparameters as follows. 
Loss weights: NLL weights $\lambda_2, \lambda_3 = 1.0$, IPM weight $\lambda_4 = 1.0$.
NF architecture: neural spline flow with 8 blocks, each having 32 hidden units and 4 hidden layers. 
Training parameters: batch size 256, learning rate 3e-4, 100 epochs.

\begin{table}
    \caption{Hyperparameters tested in a grid search for our NF framework in Section~\ref{sec:results_benchmark}.}
    \label{tab:app_mog_hyper}
    \vskip 0.15in
    \centering
    \begin{tabular}{l|l}
        Name & Values \\ \toprule
        NF mode & Chained, Triangle \\
        True transform & Linear, nonlinear \\
        Paired prop. & 0.0, 0.05, 0.1, 0.2, 0.5 \\
        Identity weight $\lambda_5$ & 1e-4, 1e-3, 1-e1, 5e-1 \\
        Paired weight $\lambda_1$ & 0.1, 1.0, 10.0, 50.0, 100.0
    \end{tabular}
    \vskip -0.1in
\end{table}

We found that results improved with paired proportion greater than zero, but with best test set MSE at paired proportion 0.1 (true transform linear) or 0.05, 0.2 (true transform nonlinear; Table~\ref{tab:app_pairedp}).
Generally, higher paired weights were better (with 10.0, 100.0, or 50.0 achieving best results depending on the NF mode and scenario).
Identity weight did not appear to have a large impact on the results for the choices studied, though we noted during experimentation that very large identity weights would restrict the map from learning anything but an identity mapping.
In comparing the best-performing Triangle and Chained NFs, we found no significant difference for a given setting of paired data or paired weight, but leave more thorough comparison of the two approaches in practice to future work.

\begin{table}    
    \caption{Performance of best NF model by proportion of paired data used during training (measured by test set MSE and test set relative NLL MSE). While some paired data improves performance over no paired data, it appears to have diminishing returns.}
    \label{tab:app_pairedp}
    \vskip 0.15in
    \begin{small}
    \centering
    \begin{tabular}{l|rrrrr}
        Paired prop. & 0.0 & 0.05 & 0.1 & 0.2 & 0.5 \\ \toprule
        Nonlin. MSE & 5.40 & 0.34 & 0.50 & 0.25 & 1.73 \\
        Linear MSE & 0.34 & $< 0.01$ &  $< 0.01$ &  $< 0.01$ & 0.01 \\
        Linear NLL & 0.02 & 0.01 & 0.01 & 0.01 & 0.01
    \end{tabular}
    \end{small}
    \vskip -0.1in
\end{table}

\subsection{OT Hyperparameters} \label{sec:app_ot_hyper}
For linear and kernel OT, we fixed argument \texttt{mu} to 0.001 and ranged argument \texttt{eta} in $[1e-4, 1e-3, 1e-2]$. For kernel OT and KGOT, we additionally ranged argument \texttt{sigma} in $[0.1, 1.0, 5.0]$. 
For KGOT, we ranged the additional regularization parameter (corresponding to the paired data) in $\texttt{alpha}$ in $[0.1, 1.0, 10.0]$.
The paired proportions and true transforms were varied as in Table~\ref{tab:app_mog_hyper}.

We found that the smaller \texttt{eta} values tended to dominate the best OT and KGOT results, while \texttt{alpha} set to 10.0 achieved the best KGOT results.
KGOT methods tended to prefer more paired data than our NF framework for the best results (with 0.5 or 0.2 being chosen).
Kernel map estimation accuracy varied widely with a clear dependence on \texttt{sigma}; in our study, larger values (5.0) were preferred.


\section{Earth-to-Mars Details}
We give details on the Earth to Mars application, including dataset details, details of our fitted NF models, and details of domain adaptation.

\subsection{Dataset and Preprocessing} \label{sec:app_e2m_data}
\begin{figure}
    \centering
    \includegraphics[width=0.5\linewidth]{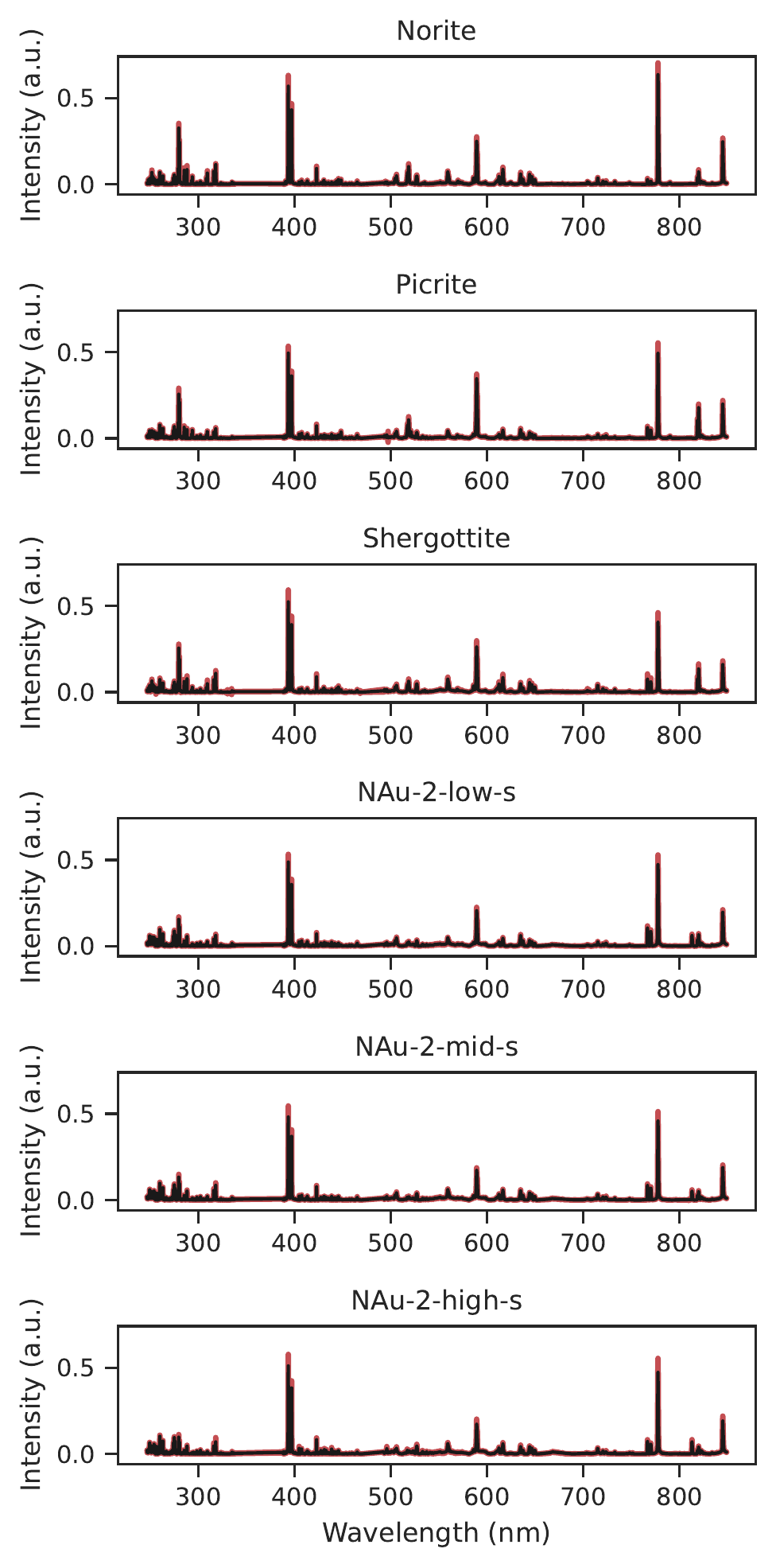}
    \caption{Mean observed spectrum (black) for each of the six Mars calibration targets, with plus or minus one standard deviation shown in red.}
    \label{fig:app_mars_targ}
\end{figure}

The ChemCam spectral measurements come from three spectrometers (UV, VIO, and VNIR ranges). 
We use data coming from wavelength ranges of [246, 338], [382, 473], and [492, 849] nanometers, respectively, resulting in a total of 5,205 measured wavelengths.
For each spectrum, we normalize the intensity values within each spectrometer to the sum of all spectral values within the spectrometer range.
Based on exploratory analyses with principal components analysis, we selected a latent dimension of 30 to represent the variation in the spectra.
The DK transform consists of a vector (the same length as a spectral measurement) that is multiplied by the spectral measurement to adjust intensity values.
The six Mars rover calibration targets are shown in Figure~\ref{fig:app_mars_targ}.

\subsection{NF Model Fitting}\label{sec:app_e2m_nf}
We used a Chained NF with a combination of a dimension-specific affine transform combined with 64 RealNVP normalizing flow blocks.
We trained for 300 epochs with a batch size of 256, a learning rate of $3e-4$ and paired weight 100.0; all other loss weights were set to 1.0.
For the sliced Wasserstein distance calculation, we used 5000 samples with 2000 projections per evaluation.
Because the paired data is comprised of multiple measurements of each target, we sampled pairs 50 times for each target to include in the training data.
We used a latent dimension of 30; we initialized a linear encoder and decoder via partial least squares on the Earth data, but allowed the weights to be updated during the NF training (see~\cref{eq:loss_gen}) to better capture variation in each data set.

\subsection{Domain Adaptation}\label{sec:app_e2m_da}
For domain adaptation, we used a multilayer perceptrons (\texttt{scikit-learn MLPRegressor}) with hidden layer sizes $[30, 30, 10]$.
We conducted a grid search (based on a 90-10 train/test split) over regularization parameter \texttt{alpha} $\in [1e-7, 1e-6, 1e-5, 1e-4, 1e-3, 1e-2, 1e-1]$ and learning rate $[1e-4, 1e-3, 1e-2]$ and used early stopping with a maximum of 5,000 epochs; other hyperparameters were set to defaults.
Because some spectral measurements are missing composition values in some oxide categories, we fit separate models for each oxide to maximize the data available for each model.
We found that all models (no DA, our DA, or domain-knowledge DA) achieved similar train and test accuracy on the labeled (possibly transformed) Earth data.
The fitted models were then applied to the Mars calibration targets to assess accuracy in predicting the known compositions.

\end{appendices}
\end{document}